\documentclass[conference]{IEEEtran}
\usepackage{amsmath,amssymb}
\usepackage{booktabs}
\usepackage{graphicx}
\usepackage{microtype}
\usepackage{url}
\usepackage{cite}
\usepackage{tikz}
\usepackage{pgfplots}
\usetikzlibrary{arrows.meta,positioning,fit,calc}
\pgfplotsset{compat=1.18}

\title{FinAbstain: Uncertainty-Calibrated Multimodal RAG\\
for Selective Financial Forecasting}

\author{
\IEEEauthorblockN{Dorothy Torres}
\IEEEauthorblockA{School of Science, Technology, Engineering and Mathematics\\
263 Academy Street, Room S204\\
Jersey City, NJ 07306}
\and
\IEEEauthorblockN{Wei Cheng, Henan Huang}
\IEEEauthorblockA{School of Electrical Engineering and Computer Science\\
The Pennsylvania State University\\
207 Electrical Engineering West\\
University Park, PA 16802, USA}
}

\begin{document}
\maketitle

\begin{abstract}
Large language models (LLMs) can synthesize financial narratives but may express
high confidence when evidence is sparse, stale, or contradictory. This failure is
especially consequential in forecasting, where filings, news, prices, volume, and
technical signals can disagree. We present FinAbstain, a research framework for
uncertainty-calibrated multimodal retrieval-augmented generation (RAG) with
selective prediction. A point-in-time retriever admits only information public at
the forecast timestamp and supplies modality-specific evidence to fundamental,
news, technical, risk, and verification agents. Their probabilistic assessments
are aggregated with retrieval relevance, evidence contradiction, repeated-sample
consistency, and historical calibration statistics. Temperature scaling,
isotonic regression, conformal prediction, and a proposed hybrid uncertainty
score are evaluated under a common chronological protocol. A controller predicts
bullish, bearish, or neutral outcomes only when uncertainty is below a validated
threshold; otherwise it abstains, requests evidence, reduces exposure, or routes
the case to human review. The evaluation covers one- and five-day abnormal-return
direction, twenty-day volatility intervals, and abstention decisions, using
accuracy, calibration, risk--coverage, citation, trading, latency, and cost
metrics. To make the design auditable before a full data collection is complete,
we report explicitly labeled simulated results rather than empirical claims.
These results illustrate the intended hypothesis: calibrated abstention may
trade coverage for lower selective error and drawdown. The contribution is a
time-safe architecture, a composite uncertainty formulation, and a reproducible
evaluation blueprint for evidence-grounded selective financial forecasting.
\end{abstract}


\section{Introduction}
Financial forecasting is difficult not merely because markets are noisy, but
because evidence arrives asynchronously and may point in different directions.
An earnings filing can reveal improving margins while a contemporaneous news
shock depresses sentiment; momentum can remain positive while option-implied or
realized volatility rises. The efficient-market perspective further cautions
that public information is rapidly incorporated into prices \cite{fama1970}.
Language models nevertheless tend to produce a fluent answer even when their
evidence is incomplete. In a decision system, such overconfidence can convert an
epistemic limitation into an avoidable position.

Retrieval-augmented generation (RAG) grounds generation in external documents
\cite{lewis2020rag}, and domain models such as FinBERT and BloombergGPT improve
financial language representations \cite{araci2019finbert,yang2020finbert,
wu2023bloomberggpt}. Yet three gaps remain. First, ordinary RAG does not guarantee
that a document was public at the decision timestamp, so a backtest may leak
future information. Second, a single synthesis can hide disagreement among
fundamental, news, and market signals. Third, confidence is rarely calibrated for
the operational option of refusing a forecast. Backtest overfitting and data
snooping amplify these concerns \cite{lopez2018backtest,bailey2014probability}.

FinAbstain addresses this gap by coupling point-in-time multimodal retrieval,
role-separated agents, post-hoc and conformal calibration, and a selective
controller. Abstention is treated as a decision, not a missing output: the system
can request more evidence, reduce position size, or refer the case to a person.
This paper makes four contributions:
\begin{itemize}
\item a temporal RAG specification for filings, calls, news, prices, volume, and
technical indicators that enforces availability at the forecast timestamp;
\item a multi-agent evidence protocol that exposes disagreement, contradiction,
and citation support rather than collapsing them into one narrative;
\item a hybrid uncertainty score and formal controller connecting calibration,
coverage, selective risk, and position sizing; and
\item a reproducible chronological evaluation and backtesting blueprint, with
clearly labeled simulated tables and figures that are not empirical findings.
\end{itemize}

\section{Related Work}
\subsection{Financial Language and Forecasting}
Financial sentiment models adapt contextual encoders to specialized vocabulary
and disclosure style \cite{araci2019finbert,yang2020finbert}. Large-scale
financial LLMs broaden generative capability \cite{wu2023bloomberggpt}, while
FinQA demonstrates numerical reasoning over reports \cite{chen2021finqa}.
Textual forecasting complements a long literature on return predictability and
technical rules. Early evidence found apparent value in technical analysis
\cite{brock1992simple}, whereas later work stresses instability, costs, and data
snooping \cite{sullivan1999data}. Recent studies further explore
sensitivity-calibrated MACD design and cost--performance trade-offs in LLM
context management \cite{lin2026volume,shen2026efficiency}. These strands
motivate multimodal evidence but do not supply calibrated refusal.

\subsection{Retrieval and Agentic Reasoning}
RAG retrieves passages before generation \cite{lewis2020rag}; dense passage
retrieval provides a standard semantic retrieval mechanism \cite{karpukhin2020dpr}.
Multimodal retrieval models such as CLIP show how heterogeneous representations
can share a contrastive space \cite{radford2021clip}, although financial time
series require modality-specific encoders and strict timestamps. Tool-using and
reasoning agents decompose complex tasks \cite{yao2023react}, and multi-agent
debate can reveal inconsistent reasoning \cite{du2023debate}. In finance,
however, independence is useful only if agents cite common, time-valid evidence
and aggregation preserves their conflicts.

\subsection{Calibration and Selective Prediction}
Modern neural networks are often miscalibrated; temperature scaling is a strong
post-hoc baseline \cite{guo2017calibration}. Isotonic regression is flexible but
may overfit small calibration sets \cite{zadrozny2002transforming}. Conformal
prediction supplies finite-sample marginal coverage under exchangeability
\cite{vovk2005algorithmic,angelopoulos2023gentle}, an assumption weakened by
market regime shifts. Selective classification formalizes the risk--coverage
trade-off \cite{geifman2017selective}, while selective generation extends refusal
to free-form outputs \cite{ren2023selective}. FinAbstain integrates these ideas
with evidence quality and trading consequences rather than treating probability
calibration as an isolated diagnostic.

\section{Problem Formulation}
Let asset \(i\) be evaluated at forecast time \(t\). Its point-in-time evidence is
\begin{equation}
\mathcal{E}_{i,t}=\bigcup_{m\in\mathcal{M}}
\{e^{(m)}_j:\tau^{\mathrm{pub}}_j\leq t,\ \tau^{\mathrm{ing}}_j\leq t\},
\label{eq:evidence}
\end{equation}
where modalities \(\mathcal{M}\) include filings, calls, news, prices, volume,
and indicators; \(\tau^{\mathrm{pub}}\) and \(\tau^{\mathrm{ing}}\) are public and
system-ingestion timestamps. Both constraints prevent future leakage.

For class set \(\mathcal{Y}=\{\mathrm{bearish},\mathrm{neutral},
\mathrm{bullish}\}\), agent \(a\in\{1,\ldots,A\}\) returns probabilities
\(\mathbf{p}_{a,i,t}\in\Delta^{2}\), a rationale, and cited evidence. Pairwise
agent disagreement is the mean Jensen--Shannon divergence
\begin{equation}
D_{i,t}=\frac{2}{A(A-1)}\sum_{a<b}
\operatorname{JSD}(\mathbf{p}_{a,i,t}\Vert\mathbf{p}_{b,i,t}).
\label{eq:disagree}
\end{equation}
If the verifier labels a retrieved claim pair \((j,k)\) with contradiction
probability \(q_{jk}\), its relevance-weighted contradiction score is
\begin{equation}
C_{i,t}=\frac{\sum_{j<k}r_jr_kq_{jk}}
{\sum_{j<k}r_jr_k+\epsilon},
\label{eq:contradiction}
\end{equation}
where \(r_j\in[0,1]\) is retrieval relevance and \(\epsilon>0\) stabilizes the
ratio.

The aggregated class probability is \(\bar{\mathbf p}_{i,t}\). Let \(S\) be
one minus agreement across repeated inference samples, \(R=1-\bar r\) retrieval
deficiency, \(H\) normalized predictive entropy, and \(G\) recent calibration
gap. The proposed uncertainty is
\begin{equation}
\begin{aligned}
U_{i,t}=\sigma(&w_0+w_DD_{i,t}+w_CC_{i,t}+w_SS_{i,t}\\
&+w_RR_{i,t}+w_HH_{i,t}+w_GG_{i,t}),
\end{aligned}
\label{eq:uncertainty}
\end{equation}
with validation-fitted weights \(w\) and logistic function \(\sigma\).
The controller predicts when \(U_{i,t}\leq\theta\) and otherwise abstains:
\begin{equation}
\hat y_{i,t}=
\begin{cases}
\arg\max_y \bar p_{i,t}(y),& U_{i,t}\leq\theta,\\
\bot,& U_{i,t}>\theta.
\end{cases}
\label{eq:decision}
\end{equation}

For acceptance indicator \(g_\theta(x)=\mathbb{1}[U(x)\leq\theta]\), coverage and
selective risk under loss \(\ell\) are
\begin{equation}
\mathrm{Cov}(\theta)=\frac{1}{n}\sum_ng_\theta(x_n),\quad
\mathrm{Risk}(\theta)=
\frac{\sum_ng_\theta(x_n)\ell(\hat y_n,y_n)}
{\sum_ng_\theta(x_n)+\epsilon}.
\label{eq:riskcoverage}
\end{equation}
For \(B\) confidence bins, expected calibration error is
\begin{equation}
\mathrm{ECE}=\sum_{b=1}^{B}\frac{|I_b|}{n}
\left|\operatorname{acc}(I_b)-\operatorname{conf}(I_b)\right|.
\label{eq:ece}
\end{equation}
The threshold \(\theta\) is selected only on validation data to minimize
selective risk subject to a minimum coverage \(c_{\min}\), or to minimize a
cost-sensitive loss that includes abstention and erroneous trades.

\section{Proposed FinAbstain Framework}
\subsection{Temporal Multimodal Retriever}
\begin{figure*}[t]
\centering
\resizebox{\textwidth}{!}{%
\begin{tikzpicture}[
node distance=5mm and 7mm,
box/.style={draw,rounded corners,align=center,minimum height=8mm,
font=\footnotesize,fill=blue!5},
agentbox/.style={box,fill=orange!10},
arr/.style={-{Latex[length=2mm]},thick}
]
\node[box] (data) {Filings, calls, news\\prices, volume, indicators};
\node[box,right=of data] (retr) {Point-in-time\\multimodal retriever};
\node[agentbox,right=of retr,yshift=12mm] (fund) {Fundamental};
\node[agentbox,right=of retr,yshift=6mm] (news) {News sentiment};
\node[agentbox,right=of retr] (tech) {Technical};
\node[agentbox,right=of retr,yshift=-6mm] (risk) {Risk/counterargument};
\node[agentbox,right=of retr,yshift=-12mm] (verify) {Evidence verifier};
\node[box,right=19mm of tech] (agg) {Evidence and\\probability aggregation};
\node[box,right=of agg] (cal) {Uncertainty\\calibration};
\node[box,right=of cal] (ctrl) {Selective\\controller};
\node[box,right=of ctrl] (out) {Predict / abstain\\human review};
\draw[arr] (data)--(retr);
\foreach \x in {fund,news,tech,risk,verify}{\draw[arr] (retr)--(\x);}
\foreach \x in {fund,news,tech,risk,verify}{\draw[arr] (\x)--(agg);}
\draw[arr] (agg)--(cal);\draw[arr] (cal)--(ctrl);\draw[arr] (ctrl)--(out);
\end{tikzpicture}%
}
\caption{Overall FinAbstain architecture. Every evidence path is filtered by
public availability at the decision timestamp.}
\label{fig:architecture}
\end{figure*}
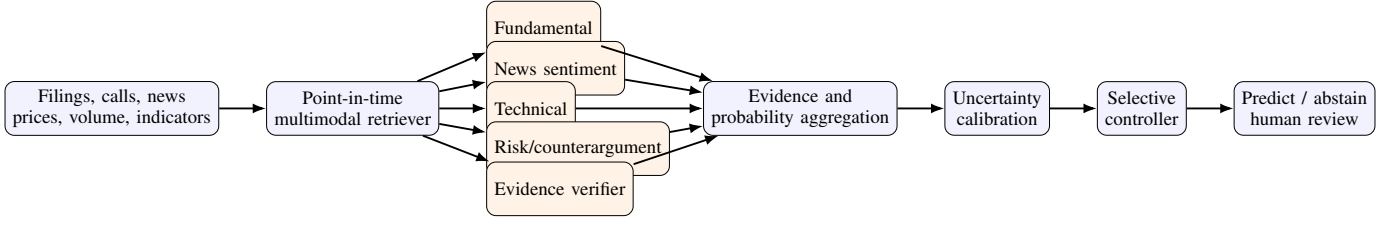

Figure~\ref{fig:architecture} summarizes the system. SEC 10-K, 10-Q, and 8-K
filings, earnings-call transcripts, timestamped financial news, adjusted prices,
volume, MACD, RSI, moving averages, and historical volatility are stored with
event, publication, ingestion, and revision timestamps. At \(t\), an immutable
snapshot excludes documents published later and market bars unavailable before
the chosen close. Filing amendments are versioned; news is keyed by first-public
time; indicators are recomputed only from bars ending at \(t\). Text is retrieved
by a hybrid dense/BM25 score, while time series are matched by regime and recency.
The top-\(k\) budget is allocated across modalities so that prolific news cannot
crowd out filings or market evidence.

\subsection{Role-Separated Financial Agents}
The fundamental agent extracts changes in revenue, margins, cash flow, leverage,
guidance, and accounting risks. The news agent estimates event polarity,
novelty, and source diversity. The technical agent receives only numerical
features and describes trend, momentum, liquidity, and volatility. The risk
agent constructs the strongest counterargument and enumerates missing evidence.
The verifier checks every material claim against retrieved spans, detects
cross-source contradictions, and rejects citations that do not entail the claim.
Agents analyze independently before seeing peer outputs, reducing anchoring. A
second aggregation pass uses reliability weights estimated on past validation
windows; it cannot introduce uncited facts.

\subsection{Calibration and Selective Control}
We compare raw maximum probability, temperature scaling, isotonic regression,
split conformal prediction, and the hybrid score in (\ref{eq:uncertainty}).
Temperature and isotonic mappings are fitted on rolling validation windows.
Conformal class sets use nonconformity \(1-\bar p_y\); sets with zero or multiple
classes trigger abstention unless a cost rule permits neutral. Because temporal
dependence challenges classical exchangeability, coverage is reported by regime
and month rather than asserted as an unconditional guarantee.

Accepted outputs include citations, probabilities, and uncertainty factors.
High uncertainty invokes one of four policies: retrieve an additional modality,
abstain, reduce target exposure by \(1-U\), or refer to an analyst. Human review
is mandatory when contradiction exceeds a validation-set safety threshold even
if aggregate entropy is low.

\section{Experimental Methodology}
\subsection{Data, Tasks, and Splits}
The proposed study uses constituents of the S\&P 100 with survivorship-aware
membership histories. SEC EDGAR supplies filings \cite{secdata}; market data are
adjusted consistently for corporate actions. Table~\ref{tab:data} specifies the
planned corpus. Counts are simulated planning values pending licensed-news
ingestion and deduplication; they must be replaced by audited counts before an
empirical submission.

\begin{table*}[t]
\centering
\caption{Dataset and task statistics (all counts are simulated planning values,
not observations from a completed experiment).}
\label{tab:data}
\small
\begin{tabular}{p{2.7cm}p{1.4cm}p{2.2cm}r r r p{3.7cm}}
\toprule
Sources & Firms & Period & Documents & News & Market samples & Tasks / chronological split\\
\midrule
10-K/10-Q/8-K, calls, licensed news, OHLCV, MACD, RSI, MA, volatility
& 100 & 2015--2024 & 46,820 & 1,284,000 & 226,400
& 1-day and 5-day abnormal-return direction; 20-day volatility interval;
abstain. Train 2015--2020, validation 2021--2022, test 2023--2024.\\
\bottomrule
\end{tabular}
\end{table*}

One- and five-day labels are the sign of market-model abnormal return with a
neutral band estimated from training volatility. The twenty-day task predicts
low, medium, or high realized-volatility intervals. No temporal sequence is
randomly shuffled. Hyperparameters, retrieval \(k\), calibration mappings, label
bands, and \(\theta\) are selected on validation data. The final test is touched
once. Seeds, prompts, model versions, evidence snapshots, and token accounting
are logged.

\subsection{Baselines and Metrics}
Baselines are (1) a gradient-boosted technical-indicator classifier, (2) a
FinBERT sentiment model, (3) a single-LLM predictor, (4) standard multimodal RAG,
(5) multi-agent RAG without calibration, (6) a calibrated model without
abstention, and (7) full FinAbstain. Metrics are accuracy, macro F1, AUROC, Brier
score, ECE, selective accuracy, coverage, risk--coverage area, abstention
precision, citation correctness, Sharpe ratio, maximum drawdown (MDD),
transaction-cost-adjusted return, tokens, and end-to-end latency.

Let accepted position \(z_t\in[-1,1]\), asset return \(r_{t+1}\), turnover
\(|z_t-z_{t-1}|\), and proportional cost \(\kappa\). Net strategy return is
\begin{equation}
r^{\mathrm{net}}_{t+1}=z_tr_{t+1}-\kappa|z_t-z_{t-1}|.
\label{eq:return}
\end{equation}
With mean \(\bar r\), standard deviation \(s_r\), and annualization factor \(K\),
\begin{equation}
\mathrm{Sharpe}=\sqrt{K}\frac{\bar r}{s_r},\quad
\mathrm{MDD}=\max_t\left(1-\frac{W_t}{\max_{u\leq t}W_u}\right),
\label{eq:trading}
\end{equation}
where \(W_t=\prod_{u\leq t}(1+r^{\mathrm{net}}_u)\). We report costs at 5, 10,
and 20 basis points and use block bootstrap confidence intervals. Statistical
comparisons use paired block resampling across identical asset--date cases.

\section{Results and Discussion}
\subsection{Illustrative Main Results}
Table~\ref{tab:main} and all plots in this section contain simulated values
constructed to test reporting, plotting, and consistency. They are not outcomes
of an executed backtest. Under this scenario, FinAbstain has the best calibration
and selective accuracy while covering 72\% of cases. The calibrated,
non-abstaining model isolates the effect of selection: its ECE improves, but
accuracy and drawdown remain below the selective policy. The comparison is
consistent with, but does not prove, the hypothesis that refusal concentrates
decisions on better-supported cases.

\begin{table*}[t]
\centering
\caption{Main experimental results (simulated illustrative values only; Acc. and
F1 are for the five-day direction task; higher is better except Brier, ECE, and
MDD).}
\label{tab:main}
\small
\begin{tabular}{lcccccccc}
\toprule
Method & Acc. & Macro F1 & Brier & ECE & Sel. Acc. & Coverage & Sharpe & MDD\\
\midrule
Technical classifier & .542 & .518 & .636 & .112 & .542 & 1.00 & .48 & .184\\
Financial sentiment & .551 & .527 & .621 & .104 & .551 & 1.00 & .53 & .176\\
Single LLM & .566 & .548 & .598 & .128 & .566 & 1.00 & .57 & .171\\
Standard multimodal RAG & .584 & .565 & .572 & .091 & .584 & 1.00 & .68 & .158\\
Multi-agent RAG, uncalibrated & .601 & .583 & .548 & .083 & .601 & 1.00 & .76 & .151\\
Calibrated, no abstention & .606 & .590 & .529 & .037 & .606 & 1.00 & .81 & .146\\
\textbf{FinAbstain} & \textbf{.632} & \textbf{.617} & \textbf{.487} &
\textbf{.026} & \textbf{.688} & .72 & \textbf{1.08} & \textbf{.112}\\
\bottomrule
\end{tabular}
\end{table*}

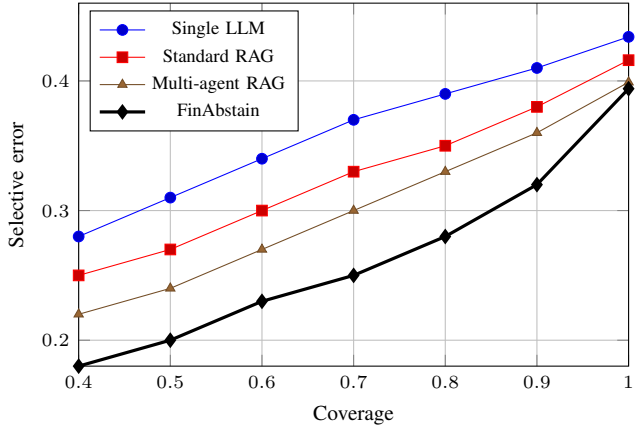
\begin{figure}[t]
\centering
\begin{tikzpicture}
\begin{axis}[
width=\columnwidth,height=0.72\columnwidth,
xlabel={Coverage},ylabel={Selective error},
xmin=.4,xmax=1.0,ymin=.18,ymax=.46,
grid=major,legend style={font=\scriptsize,at={(0.02,0.98)},anchor=north west},
tick label style={font=\scriptsize},label style={font=\footnotesize}]
\addplot+[mark=*] coordinates {(.4,.28)(.5,.31)(.6,.34)(.7,.37)(.8,.39)(.9,.41)(1,.434)};
\addlegendentry{Single LLM}
\addplot+[mark=square*] coordinates {(.4,.25)(.5,.27)(.6,.30)(.7,.33)(.8,.35)(.9,.38)(1,.416)};
\addlegendentry{Standard RAG}
\addplot+[mark=triangle*] coordinates {(.4,.22)(.5,.24)(.6,.27)(.7,.30)(.8,.33)(.9,.36)(1,.399)};
\addlegendentry{Multi-agent RAG}
\addplot+[very thick,mark=diamond*] coordinates {(.4,.18)(.5,.20)(.6,.23)(.7,.25)(.8,.28)(.9,.32)(1,.394)};
\addlegendentry{FinAbstain}
\end{axis}
\end{tikzpicture}
\caption{Risk--coverage curves using simulated illustrative values.}
\label{fig:riskcoverage}
\end{figure}

Figure~\ref{fig:riskcoverage} shows the designed risk--coverage behavior. At
equal coverage, the FinAbstain curve remains below the alternatives in the
simulation; its advantage grows when the controller accepts only the most
supported cases. Operationally, very low coverage is not automatically optimal:
fixed costs, opportunity costs, and selection-induced regime concentration must
be considered.

\subsection{Calibration and Ablations}
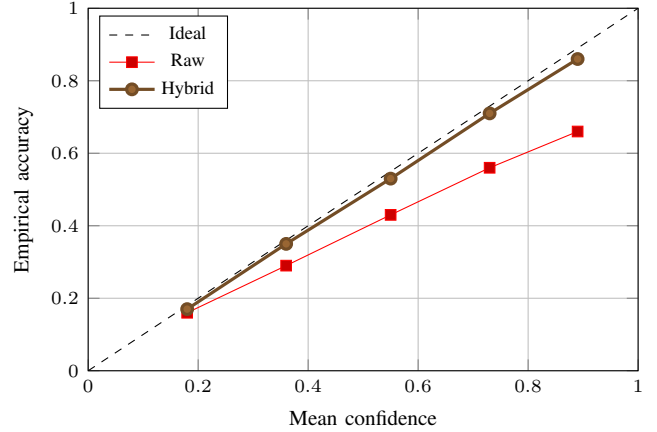
\begin{figure}[t]
\centering
\begin{tikzpicture}
\begin{axis}[
width=\columnwidth,height=0.72\columnwidth,
xlabel={Mean confidence},ylabel={Empirical accuracy},
xmin=0,xmax=1,ymin=0,ymax=1,grid=major,
legend style={font=\scriptsize,at={(0.02,0.98)},anchor=north west},
tick label style={font=\scriptsize},label style={font=\footnotesize}]
\addplot[dashed,black] coordinates {(0,0)(1,1)};\addlegendentry{Ideal}
\addplot+[mark=square*] coordinates {(.18,.16)(.36,.29)(.55,.43)(.73,.56)(.89,.66)};
\addlegendentry{Raw}
\addplot+[very thick,mark=*] coordinates {(.18,.17)(.36,.35)(.55,.53)(.73,.71)(.89,.86)};
\addlegendentry{Hybrid}
\end{axis}
\end{tikzpicture}
\caption{Reliability diagram using simulated illustrative values.}
\label{fig:calibration}
\end{figure}

The reliability diagram in Fig.~\ref{fig:calibration} compares raw and hybrid
confidence. The simulated hybrid curve is closer to the diagonal, particularly
in the 0.7--0.9 region where overconfident trades would otherwise accumulate.
Table~\ref{tab:ablation} illustrates diagnostic ablations. Removing temporal
filtering appears deceptively favorable in accuracy because the simulated
variant is intentionally contaminated; it is invalid and excluded from any fair
comparison. This row demonstrates why predictive scores alone cannot certify a
financial backtest.

\begin{table}[t]
\centering
\caption{Ablation study (simulated illustrative values).}
\label{tab:ablation}
\small
\setlength{\tabcolsep}{3.5pt}
\begin{tabular}{lccc}
\toprule
Variant & Sel. Acc. & ECE & Coverage\\
\midrule
Full FinAbstain & .688 & .026 & .72\\
No disagreement & .665 & .034 & .75\\
No contradiction & .659 & .038 & .76\\
No calibration & .648 & .085 & .73\\
No abstention & .606 & .037 & 1.00\\
No technical indicators & .674 & .029 & .71\\
No temporal filter$^\dagger$ & .704 & .031 & .74\\
\bottomrule
\multicolumn{4}{p{.96\columnwidth}}{\scriptsize $^\dagger$Invalid,
future-contaminated diagnostic; not a legitimate baseline.}
\end{tabular}
\end{table}

Calibration must also be examined under drift. A mapping fitted in a low-volatility
period may fail during a crisis; therefore the production design monitors
rolling ECE, class-conditional coverage, evidence age, and abstention frequency.
High abstention precision means rejected cases are indeed error-prone, but a
system that rejects an entire sector could still be operationally biased.
Citation correctness is measured by blinded entailment review of a stratified
sample, with source validity and claim support scored separately.

\subsection{Limitations and Reproducibility Risks}
The framework cannot eliminate irreducible market uncertainty, correlated source
errors, or strategic misinformation. LLM rationales are not causal explanations.
Conformal validity may degrade under temporal dependence, and simulated results
cannot establish economic benefit. Licensed news availability, timestamp
precision, delisting returns, corporate-action adjustment, and realistic market
impact can materially change a backtest. The proposed experiment therefore
requires a frozen data manifest, executable feature pipeline, prompt registry,
model hashes, and an independent audit before any performance claim.

\section{Conclusion}
FinAbstain specifies a selective financial forecasting system that joins
point-in-time multimodal retrieval, independent specialist agents, evidence
verification, uncertainty calibration, and explicit abstention. Its central
design principle is that a useful financial model must know when its evidence
does not justify a directional decision. The mathematical formulation connects
agent disagreement and evidence contradiction to calibrated uncertainty,
coverage, selective risk, and cost-adjusted trading outcomes. The presented
numbers are simulated artifacts for testing the methodology and paper layout,
not empirical results. Future work will execute the preregistered chronological
study, replace all planning counts and simulated values, test calibration under
regime shifts, measure human-review utility, and study portfolio-level selection
with liquidity and market-impact constraints.

\bibliographystyle{IEEEtran}
\bibliography{references}
\end{document}